\documentclass[review,12pt]{elsarticle}

\usepackage[T1]{fontenc}
\usepackage[utf8]{inputenc}
\usepackage{amsmath,amssymb}
\usepackage{graphicx}
\usepackage{booktabs}
\usepackage{hyperref}
\usepackage{xurl} 
\usepackage{url}
\usepackage{natbib}
\usepackage{lineno}
\usepackage{xcolor}
\usepackage{microtype}
\usepackage{cleveref}
\usepackage{float}
\usepackage{rotating}

\makeatletter
\def\ps@pprintTitle{%
  \let\@oddhead\@empty
  \let\@evenhead\@empty
  \def\@oddfoot{\footnotesize\itshape Garc\'ia Crespi et al.\hfill\thepage}%
  \def\@evenfoot{\footnotesize\itshape Garc\'ia Crespi et al.\hfill\thepage}%
}
\makeatother
\journal{International Journal of Forecasting}

\begin{document}

\begin{frontmatter}

\title{Rolling-Origin Validation Reverses Model Rankings in Multi-Step PM10 Forecasting: XGBoost, SARIMA, and Persistence}

\author[umhcomp]{Federico Garc\'ia Crespi\corref{cor1}}
\ead{fedeg@umh.es}

\author[umhfis]{Eduardo Yubero Funes}
\author[umhfis]{Marina Alfosea Sim\'on}

\address[umhcomp]{Departamento de Ingenier\'ia de Computadores, Universidad Miguel Hern\'andez, Elche, Spain}
\address[umhfis]{Departamento de F\'isica Aplicada, Universidad Miguel Hern\'andez, Elche, Spain}

\begin{abstract}
(a) Many air quality forecasting studies report gains from machine learning, but evaluations often use static chronological splits and omit persistence baselines, so the operational added value under routine updating is unclear.

(b) Using 2,350 daily PM10 observations from 2017 to 2024 at an urban background monitoring station in southern Europe, we compare XGBoost and SARIMA against persistence under a static split and a rolling-origin protocol with monthly updates. We report horizon-specific skill and the predictability horizon, defined as the maximum horizon with positive persistence-relative skill. Static evaluation suggests XGBoost performs well from one to seven days ahead, but rolling-origin evaluation reverses rankings: XGBoost is not consistently better than persistence at short and intermediate horizons, whereas SARIMA remains positively skilled across the full range.

(c) For researchers, static splits can overstate operational usefulness and change rankings. For practitioners, rolling-origin, persistence-referenced skill profiles show which methods stay reliable at each lead time.
\end{abstract}


\end{frontmatter}


\section{Introduction}
\label{sec:introduction}

Air quality forecasting is operationally useful only if predictions remain informative at the time scales required for intervention. For PM$_{10}$, this requirement is particularly stringent. Under Directive 2008/50/EC, Member States must limit daily exceedances of the 24-hour mean concentration threshold of 50\,$\mu$g\,m$^{-3}$, making short-term forecasting relevant for episode management, traffic regulation, industrial mitigation, and public health advisories \citep{EuropeanParliamentandCouncil2008,Krylova2021,Vlachogianni2011}. In this context, the practical question is not merely whether a model reduces average forecast error, but whether it provides actionable skill beyond simple temporal inertia over the one- to seven-day horizon relevant to operational decision-making.

A large body of literature has addressed PM forecasting using statistical, machine learning, and deep learning approaches. Despite this methodological diversity, the operational meaning of reported gains remains difficult to assess because evaluation is often based on a single static train--test split \cite{tashman2000} and summarized through aggregate error measures such as RMSE, MAE, or $R^2$. These metrics are informative but incomplete: they do not show whether a model consistently improves upon a naive persistence forecast, nor how forecast usefulness changes across the multi-step horizon. This limitation is especially important in highly autocorrelated environmental series, where strong short-lead performance may reflect temporal inertia more than genuine predictive skill. As a result, model rankings obtained under static evaluation may be misleading from an operational perspective, particularly when forecasting systems are used under sequential information arrival and repeated forecast origins. In this setting, temporal validation design is not a secondary implementation detail but part of the substantive forecasting question.

Recent work has also examined direct and recursive multi-step PM$_{10}$ forecasting with deep neural architectures in Southern Spain, showing that convolutional models can perform well for short-horizon prediction and that recursive forecasting may remain usable only under limited background conditions \citep{GomezGomez2025}. However, this line of work remains primarily architecture-centered and accuracy-centered. It does not directly address whether comparative conclusions are stable under leakage-safe rolling-origin evaluation, whether apparent gains persist relative to a persistence benchmark, or how long predictive usefulness remains operationally meaningful across horizons.

A second limitation is that most comparative studies focus on performance at one predefined horizon or report multi-step errors without identifying the horizon beyond which predictive skill becomes negligible. Yet for operational forecasting, horizon dependence is central. Forecast usefulness should be assessed not only by absolute error but also by how long a model preserves positive skill relative to a minimal benchmark. Persistence is especially relevant in this setting because it represents the simplest plausible forecast in short-memory pollution series. Evaluating models against persistence makes it possible to distinguish genuine predictive value from improvements that are statistically detectable but operationally trivial.

This study addresses these issues by reframing multi-step PM$_{10}$ forecasting as a problem of operational predictability under realistic temporal evaluation. Using daily PM$_{10}$ series from an urban background station in southeastern Spain, we compare three forecast families with different complexity profiles: persistence, SARIMA, and XGBoost. The comparison is conducted under a leakage-safe rolling-origin evaluation with train-only preprocessing, so that model assessment remains aligned with the information actually available at forecast time.

This paper makes three main contributions. First, it provides a reproducible evaluation design for multi-step PM$_{10}$ forecasting that distinguishes genuine forecast skill from artifacts introduced by static validation and non-causal preprocessing. Second, it introduces an operational interpretation of forecast usefulness through the predictability horizon, H$^*$, defined as the maximum horizon at which a model retains positive persistence-relative skill. Third, it empirically shows that model rankings are not invariant to evaluation design: under rolling-origin evaluation, XGBoost — which appeared uniformly superior under static evaluation — failed to systematically outperform persistence at short and intermediate horizons, whereas SARIMA maintained positive skill throughout, reversing the apparent ranking. This ranking sensitivity is not a marginal technical detail, but a substantive result for forecasting methodology and practice.

The remainder of the paper is organized as follows. Section~\ref{sec:methodology} describes the dataset, forecasting protocols, models, and evaluation metrics. Section~\ref{sec:elche_results} presents the empirical case study results. Section~\ref{sec:discussion} interprets the findings from an operational forecasting perspective, with emphasis on persistence-relative skill and H$^*$. Section~\ref{sec:conclusion} concludes with implications for PM$_{10}$ forecasting and the broader evaluation of environmental time-series models.

\section{Methodology}
\label{sec:methodology}

This study evaluates multi-step operational PM$_{10}$ forecasting under temporally realistic validation protocols. The objective is not only to compare predictive accuracy between models but also to determine the forecast horizon at which each model retains useful skill relative to a persistence benchmark. To this end, the experimental design combines daily PM$_{10}$ observations, competing forecasting models with different complexity profiles, and temporal evaluation schemes that distinguish static holdout assessment from rolling-origin deployment conditions.

\subsection{Study Design}
\label{sec:studydesign}

The empirical design is structured around three methodological questions. First, how do forecast rankings change when evaluation moves from a conventional static split to rolling-origin evaluation? Second, to what extent do more flexible models improve upon a simple persistence benchmark across lead times from one to seven days ahead? Third, at which horizon does the model skill become operationally negligible?

To answer these questions, we compare three forecast families: persistence, SARIMA, and XGBoost. The comparison is conducted under alternative temporal validation schemes and is evaluated using both absolute error metrics and persistence-relative skill. This design allows for the separation of raw predictive performance from operational usefulness under realistic forecasting conditions.

\subsection{Study Area and Data}
\label{sec:data}

The empirical illustration uses daily PM$_{10}$ concentration data from an urban background monitoring station in Elche (Alicante, southeastern Spain). This site is representative of a Mediterranean urban environment where PM$_{10}$ dynamics are influenced by local emissions, synoptic variability, resuspension processes, and episodic regional transport. Such conditions make the station a suitable test case for multi-step forecasting under heterogeneous and non-stationary environmental forcing.

The target variable is the daily mean PM$_{10}$ concentration. Forecast experiments are defined for horizons $h = 1, \dots, 7$ days ahead. The modeling pipeline uses only information available up to the forecast origin for each prediction so that the evaluation remains consistent with operational deployment. Any preprocessing, feature construction, or model fitting step is constrained by temporal ordering to prevent leakage from future observations.

\subsection{Forecasting Protocols}
\label{sec:protocols}

Two temporal evaluation settings are considered.

\textbf{Static chronological split.} In the static setting, the series is divided into a single training segment and a subsequent test segment. Models are estimated once on the training data and then evaluated during the holdout test period. This design is common in environmental forecasting studies because of its simplicity, but it does not reproduce the repeated updating process that characterizes operational forecasting systems; more broadly, the choice of performance estimation method can materially affect conclusions and model rankings \cite{cerqueira2020}.

\textbf{Rolling-origin evaluation.} In the rolling-origin setting \cite{hyndman2021}, the forecast origin advances sequentially over time. At each origin, the model is trained using only the data available up to that point, and a multi-step forecast is generated for horizons $h = 1, \dots, 7$. The procedure is then repeated at the next origin. This protocol more faithfully represents real deployment because it preserves temporal causality and allows model estimation to adapt as new observations become available.

When preprocessing is required, we use train-only preprocessing: all transformations are estimated using training data only at each forecast origin and then applied to the corresponding forecast instance. This constraint is essential to ensure that reported performance reflects the information genuinely available at prediction time. In the empirical implementation, the initial training window spans 2017--2019, the rolling-origin evaluation covers 2020--2023 with monthly fold updates, and the SARIMA order is identified once in the initial window and held fixed across all folds.

\subsection{Forecasting Models}
\label{sec:models}

Three forecasting approaches are examined.

\textbf{Persistence.} Persistence is the baseline benchmark. For each forecast origin, it assumes that the future PM$_{10}$ values remain equal to the most recently observed concentration. Although simple, persistence is a demanding comparator in highly autocorrelated environmental series and therefore provides a meaningful operational reference.

\textbf{SARIMA.} Seasonal autoregressive integrated moving average modelling represents the classical statistical family in the comparison. SARIMA captures the linear autoregressive structure, differencing dynamics, and seasonal dependence, providing an interpretable benchmark for structured time-series forecasting. Its inclusion allows for the assessment of whether more flexible methods outperform a well-established parametric alternative rather than only a naive baseline.

\textbf{XGBoost.} Extreme Gradient Boosting represents the machine learning family. XGBoost is included because it can model nonlinear relationships and interactions while remaining robust in tabular forecasting settings. In the present study, it serves as a flexible alternative to linear time-series models and enables the evaluation of whether greater functional capacity translates into durable, persistence-relative skill across horizons.

These three models were selected to span increasing levels of structural flexibility: naive temporal inertia (persistence), classical stochastic structure (SARIMA), and nonlinear machine learning (XGBoost).

\subsection{Forecast Horizons}
\label{sec:horizons}

The study focuses on direct operational relevance by evaluating forecasts from one to seven days ahead. This horizon range matches the time window in which short-term air quality forecasts are most useful for the anticipation of episodes, mitigation planning, and public communication. Assessing performance horizon by horizon is necessary because environmental predictability typically degrades with lead time, and aggregate summaries can obscure the point at which forecast skill becomes too small to support action.

\subsection{Evaluation Metrics}
\label{sec:metrics}

Model performance is assessed using standard forecast accuracy metrics together with persistence-relative skill measures.

Absolute performance is summarized using conventional error metrics, including root mean square error (RMSE) and mean absolute error (MAE). These metrics quantify forecast deviations on the original measurement scale and remain useful for comparing raw predictive accuracy across models and horizons.

However, absolute error alone is insufficient for operational interpretation. A model may improve RMSE or MAE only marginally while offering no meaningful advantage over persistence. For this reason, the central evaluative criterion in this study is persistence-relative skill. Let $\mathrm{Err}_{m}(h)$ denote the forecast error of model $m$ on the horizon $h$, and let $\mathrm{Err}_{\mathrm{pers}}(h)$ denote the corresponding persistence error. Persistence-relative skill is defined as

\begin{equation}
SS_m(h) = 1 - \frac{\mathrm{Err}_{m}(h)}{\mathrm{Err}_{\mathrm{pers}}(h)}.
\end{equation}

A positive value indicates an improvement over persistence, a zero value indicates parity with persistence, and a negative value indicates that the model performs worse than persistence. This normalization enables comparisons that are operationally interpretable across horizons.

\subsection{Definition of the Predictability Horizon, \texorpdfstring{H\textsuperscript{*}}{H*}}
\label{sec:hstar}

To summarize the usefulness of the horizon-dependent forecast, we define the predictability horizon, H$^*$, as the maximum forecast horizon at which a model preserves positive persistence-relative skill:

\begin{equation}
H^* = \max \left\{ h \in \{1,\dots,7\} : SS_m(h) > 0 \right\}.
\end{equation}

Under this definition, H$^*$ is not a claim of theoretical predictability in the information-theoretic sense, but rather an operational criterion tied to benchmark-relative usefulness. A model with larger H$^*$ remains informative over a longer decision horizon; a model with H$^*=0$ provides no advantage over persistence at any evaluated lead time.

This formulation has two practical advantages. First, it converts a vector of horizon-specific skill values into a single interpretable summary of forecast durability. Second, it prevents overinterpretation of isolated short-horizon gains by requiring usefulness to be judged using persistence-relative skill at each lead time. In this way, H$^*$ complements the standard accuracy metrics and makes the evaluation directly relevant to operational forecasting decisions.

\subsection{Comparative Logic}
\label{sec:comparativelogic}

The comparison is interpreted along two axes. The first is \textit{accuracy}, assessed through horizon-specific error values. The second is \textit{operational usefulness}, assessed through persistence-relative skill and H$^*$. This distinction is important because the model with the lowest average error is not necessarily the model with the most durable predictive usefulness under realistic deployment conditions.

Accordingly, the main empirical result sought in this study is not simply whether one model is more accurate than another, but whether model rankings remain stable once the evaluation is conducted under rolling-origin conditions and interpreted using persistence-relative skill. Any reduction, disappearance, or reversal of apparent advantages across horizons is treated as substantively informative evidence about operational predictability rather than as a secondary robustness check.

\section{Results}
\label{sec:elche_results}

\subsection{Case Study Results: Elche Urban Background Station}

This section reports the empirical results obtained for the Elche urban background station under the evaluation protocols defined in Section~\ref{sec:methodology}. Results are presented by forecast horizon and model family in order to compare absolute error, persistence-relative skill, and ranking stability under static and rolling-origin evaluation.

\subsubsection{Data Characteristics}

\begin{figure}[htbp]
\centering
\includegraphics[width=0.95\textwidth]{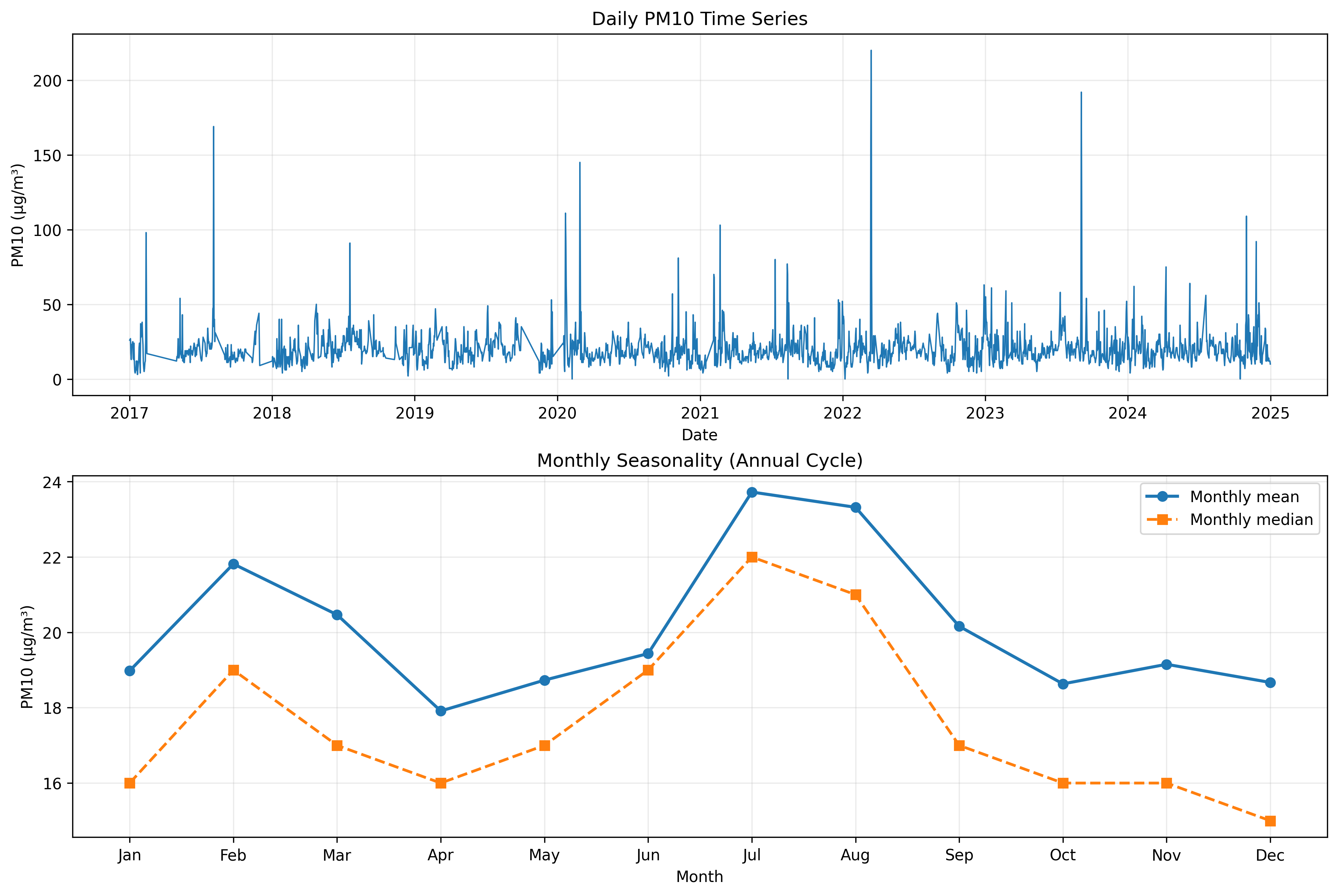}
\caption{Two-panel diagnostic summary of the daily PM10 series used in the empirical case study. Panel (a) shows the full daily time series. Panel (b) summarizes monthly seasonality through the distribution of monthly mean and median concentrations across the annual cycle.}
\label{fig:pm10_diagnostics}
\end{figure}

Quality-filtered data comprised 2,350 daily observations (January 2017--December 2024), representing
65.1\% of raw records (2,920). The series exhibits a mean PM$_{10}$ concentration of 20.1~$\mu$g\,m$^{-3}$ (median 18.0~$\mu$g\,m$^{-3}$), with a right-skewed distribution reflecting episodic Saharan dust intrusions (maximum: 218.3~$\mu$g\,m$^{-3}$). Annual data coverage improved from 51.2\% in 2017 to more than
92\% from 2020 onward, with 2023 test-year coverage at 94.2\% (344 observations). Exceedances of the
daily limit value of 50\,$\mu$g\,m$^{-3}$ constituted 2.5\% of the observations in the 2023 test year,
consistent with the predominantly Mediterranean urban-background character of the site.
Figure~\ref{fig:pm10_diagnostics} summarizes the temporal series and annual cycle of the daily PM10 series used in the empirical case study.

\subsubsection{Static Chronological Split: Apparent Skill Stability}

We first evaluated XGBoost under a static chronological split using 1,645 training observations
(2017--2022) and 325 held-out test observations from 2023 after feature engineering and horizon
alignment. Forecast skill was computed against a standard persistence baseline for horizons
$h=1,\ldots,7$.

Under this protocol, XGBoost achieved positive skill across all horizons, with
$\mathrm{SS}=0.231$--$0.299$, yielding a nominal $H^*=7$. The profile was not strictly monotonic,
with a local minimum at $h=3$, but it remained positive throughout. On this basis alone, the model would
appear to provide stable multi-step added value relative to persistence. However, this conclusion
changes materially once the validation protocol is made sequential and deployment-realistic.

\subsubsection{Rolling-Origin Evaluation: XGBoost}

We then re-evaluated XGBoost under rolling-origin evaluation with monthly fold updates over
2020--2023, using an expanding training window and train-only preprocessing within each fold.
Forty-seven valid monthly folds were retained after applying the minimum test-size criterion.

Under rolling-origin evaluation, the apparent superiority of XGBoost no longer held uniformly.
Mean persistence-relative skill was negative at $h=1$ ($\mathrm{SS}=-0.192$) and $h=3$
($\mathrm{SS}=-0.022$), near zero at $h=2$ and $h=4$, and positive only at longer horizons
($h=5$--7, $\mathrm{SS}=0.067$--$0.137$). The short-horizon failure was especially clear at
$h=1$, where 34 of 47 folds yielded non-positive skill, and the median skill was negative.

This result indicates that the static protocol materially overstated the operational utility of
XGBoost. In particular, the same model that appeared uniformly superior to persistence under a single
held-out split did not systematically beat persistence once evaluated under sequentially updated,
deployment-realistic conditions.

\subsubsection{Rolling-Origin Evaluation: SARIMA}

We next evaluated a seasonal statistical baseline under the same rolling-origin evaluation.
SARIMA order was identified once on the initial 2017--2019 training window and then applied under
monthly rolling-origin evaluation with fixed parameters and within-fold state updating.
The selected specification was SARIMA$(2,0,2)(0,1,1)_7$.

Unlike XGBoost, SARIMA maintained a positive mean skill across the full horizon range.
Mean skill values were $\mathrm{SS}=0.027$ at $h=1$, increasing to
$\mathrm{SS}=0.203$ at $h=6$ and remaining positive at $h=7$ ($\mathrm{SS}=0.192$).
Thus, SARIMA achieved a nominal $H^*=7$ while also exhibiting a more robust short-horizon profile
than XGBoost under the same temporal protocol.

Most importantly, SARIMA outperformed XGBoost at every forecast horizon under rolling-origin
evaluation. This reversal of model ranking relative to the static-split picture is the central
empirical result of the Elche case study.

\subsubsection{Comparative Interpretation}

Taken together, the three experiments show that validation design can materially alter both the
magnitude of apparent forecast skill and the ranking between competing model classes. Under static
evaluation, XGBoost appeared uniformly superior to persistence and supported a straightforward
$H^*=7$ interpretation. Under rolling-origin evaluation, this picture changed substantially:
XGBoost no longer maintained consistent short-horizon superiority, whereas SARIMA remained positive
throughout and dominated XGBoost across $h=1$--7.

The practical implication is that model complexity alone is not a reliable guide to operational
forecasting value. In this case study, a classical seasonal statistical model proved to be more robust than
a more flexible machine-learning alternative once evaluation was aligned with deployment conditions.
Accordingly, $H^*$ should not be interpreted in isolation. Its operational meaning depends on the
full horizon-wise skill profile and on the temporal validation protocol under which it is computed.

\begin{figure}[htbp]
\centering
\includegraphics[width=0.92\textwidth]{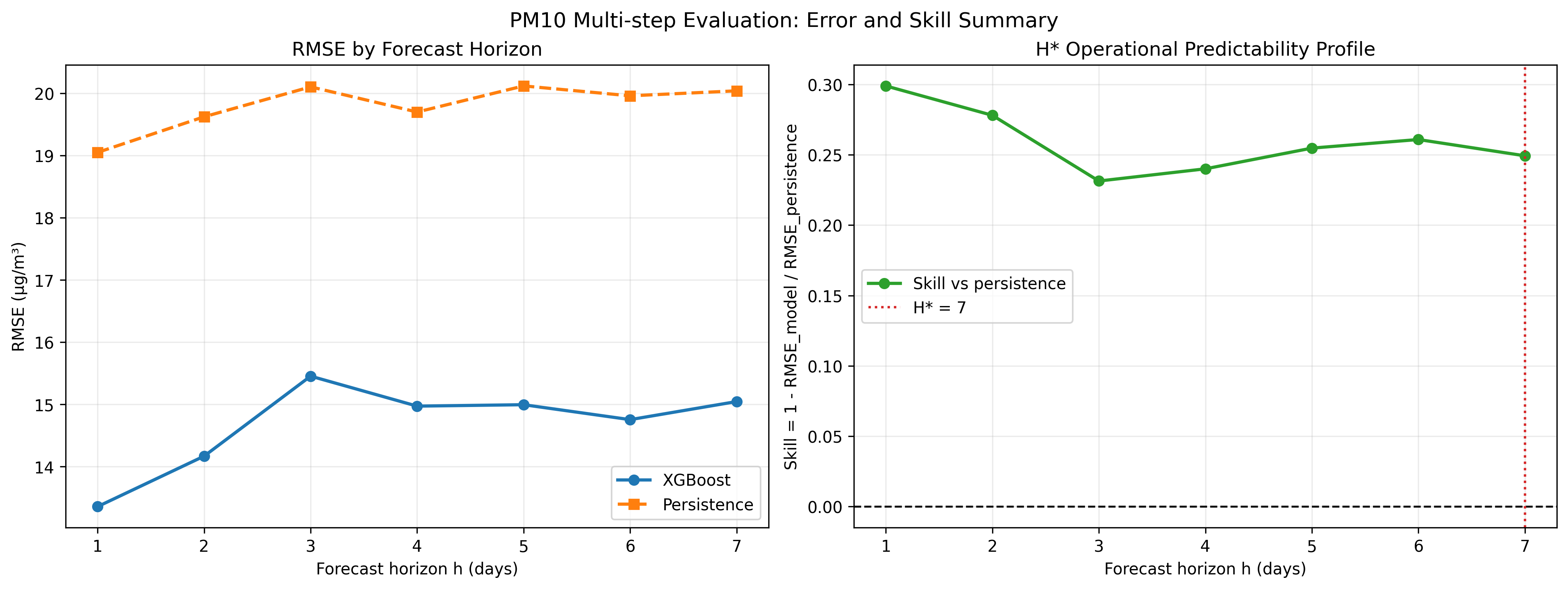}
\caption{Static-split H$^*$ evaluation for XGBoost on the empirical case study. Under a single chronological train/test partition, XGBoost remains above persistence across all horizons ($\mathrm{SS}=0.231$--$0.299$), yielding a nominal $H^*=7$. This result is informative as a single-split benchmark, but it is not representative of the rolling-origin ranking reported in the main text.}
\label{fig:hstar_real}
\end{figure}

\section{Discussion}
\label{sec:discussion}

\subsection{Validation Protocol Reverses Model Rankings}
\label{sec:rankingsreversal}

These results have two methodological implications for multi-step air-quality forecasting. First, they show that model assessment cannot be separated from temporal validation design. Apparent gains observed under a single static split may not reflect stable forecasting skill once performance is examined across repeated forecast origins and under information constraints consistent with operational use. In this sense, evaluation design is not merely a reporting choice, but part of the substantive inferential framework through which model usefulness is established.

Second, the results reinforce the importance of interpreting forecast performance relative to a minimal operational benchmark and across the full horizon of interest. In highly autocorrelated pollution series, improvements in aggregate error do not necessarily imply meaningful added value in practice. A persistence-relative perspective, together with the predictability horizon H$^*$, makes it possible to distinguish between nominal predictive accuracy and sustained operational usefulness. This distinction is especially relevant when forecast outputs are intended to support intervention timing rather than retrospective model comparison alone.

\subsection{\texorpdfstring{H$^*$ Interpretation}{H* Interpretation}: Beyond the Nominal Value}
\label{sec:hstarbeyond}

The Elche case study highlights that identical H$^*$ values may conceal materially different operational profiles. In this setting, models with similar nominal predictability horizons differed substantially in the consistency of their persistence-relative skill across intermediate lead times, indicating that horizon-specific behavior remains important even when the terminal horizon summary is unchanged.

This ranking sensitivity has an important methodological implication: conclusions about model adequacy in multi-step forecasting depend not only on model class but also on whether evaluation preserves the temporal structure of deployment. In practice, this means that apparent superiority under a single holdout period should be interpreted cautiously unless it remains stable across repeated forecast origins and against a transparent operational benchmark.

This contrast also suggests that H$^*$ should not be interpreted in isolation. For operational assessment, it is informative to examine the full skill profile $\text{Skill}(h)$ across the horizon range, since models with similar terminal horizons may differ substantially in the stability and consistency of their persistence-relative performance across intermediate lead times. From this perspective, H$^*$ is most useful when read together with the horizon-wise skill trajectory rather than as a stand-alone summary.

\subsection{Implications for Air Quality Management}
\label{sec:implications}

\textbf{Deciding when an advanced model is worth deploying.} The H$^*$ metric answers a central operational question relative to a specified baseline (here, persistence) under an explicit temporal validation protocol: it summarizes the nominal upper horizon at which mean skill is positive, and the accompanying skill profile quantifies the magnitude and fold-to-fold stability of any gains. An agency can benchmark a candidate model against persistence under rolling-origin evaluation using $H^*$ together with the full horizon-wise skill profile, thereby grounding deployment decisions in protocol-consistent evidence rather than in a single static split that can overstate apparent operational utility.

\textbf{Horizon-specific reliability for warnings and episode management.} The skill profile
underlying H$^*$ enables horizon-specific decision rules. For the one-to-seven-day forecast window considered in this study, agencies can identify horizon-specific operational regimes rather than assume uniform model superiority: in our case study, XGBoost did not systematically outperform persistence at short and intermediate horizons, whereas SARIMA maintained positive mean skill throughout the full range.

\textbf{Single-station scope and generalizability.} The empirical illustration in this study uses a single monitoring station (Elche, southeastern Spain). This setting serves to illustrate the evaluation logic of the H$^*$ framework and to show how leakage-safe temporal validation can alter conclusions about operational skill in practice. However, the empirical patterns reported here should be interpreted as station-specific evidence rather than as directly generalizable across sites, and the extension to multi-station settings remains an important next step.

\subsection{Limitations}
\label{sec:limitations}

A further limitation concerns the conditional interpretation of $H^*$ itself. First, the persistence benchmark used in this study is operationally appropriate as a transparent reference, but it is not necessarily the strongest naive baseline at longer horizons; seasonal or climatological baselines may become increasingly competitive as the horizon increases, and future work should evaluate $H^*$ relative to multiple baseline families rather than persistence alone. Second, the present analysis is descriptive rather than inferential: positive mean skill was not subjected to formal significance testing under serial dependence here, so $H^*$ should not be interpreted as a hypothesis-test threshold but as an empirical summary conditional on the observed folds. Third, continuous-error metrics such as RMSE and skill scores do not fully capture operational utility for rare exceedance events; a model may perform modestly in continuous terms yet still be valuable for threshold-oriented warning tasks, so future work should complement regression skill with exceedance-focused evaluation. Fourth, the reported predictability horizon is conditional on the information set used here, which is intentionally restricted to local PM$_{10}$ history and associated covariates; they should therefore be interpreted as operational limits under partial information, not as absolute physical predictability limits of the atmosphere. Finally, $H^*$ is inherently dependent on spatial and temporal aggregation: the horizon obtained from daily station-level PM$_{10}$ cannot be assumed to transfer unchanged to hourly, weekly, multi-station, or spatially aggregated settings.

\section{Conclusion}
\label{sec:conclusion}

This study examined multi-step PM$_{10}$ forecasting from the perspective of operational predictability under temporally coherent evaluation. The main methodological message is that conclusions about forecast usefulness depend not only on model class or aggregate error magnitude, but also on whether evaluation preserves temporal order, avoids non-causal preprocessing, and assesses skill relative to an operationally meaningful benchmark. From this perspective, validation design is part of the inferential basis on which forecasting claims should be interpreted.

The Elche case study supports three broad conclusions. First, persistence provides a strong and necessary reference for interpreting whether apparent predictive gains are substantively useful in practice. Second, model adequacy is horizon-dependent and may change once performance is examined across repeated forecast origins rather than under a single static split. Third, the predictability horizon H$^*$ is most informative when interpreted together with the full horizon-wise skill profile since operational usefulness depends not only on whether skill remains positive but also on how consistently it is sustained across lead times.

These findings also suggest a corresponding set of methodological priorities for multi-step air-quality forecasting studies. Forecast evaluation should use rolling-origin evaluation with train-only preprocessing so that model assessment remains aligned with the information actually available at forecast time. Persistence-relative skill should be reported alongside conventional error metrics, since lower RMSE or MAE does not necessarily imply meaningful operational improvement. In addition, H$^*$ should be interpreted as a protocol-dependent operational summary rather than as an intrinsic property of the pollutant series or the model class. Its value lies precisely in making explicit how forecast usefulness depends on evaluation design.

The proposed framework is intended to be model-agnostic and transferable in methodological terms. Although the present illustration is based on a single Mediterranean urban background station, the evaluation logic extends naturally to other monitoring sites, pollutants, and forecasting architectures, provided that temporal causality is preserved and skill is interpreted as persistence-relative skill. What should transfer across settings is not a fixed numerical horizon but the discipline of evaluating whether forecast usefulness survives under deployment-realistic conditions.

Several limitations should be stated explicitly. The empirical analysis is based on one station and a restricted set of forecast families, so the reported H$^*$ values should not be generalized beyond the present setting. In addition, H$^*$ is a compact summary and does not replace the full horizon-wise skill profile, which remains necessary for interpretation when models exhibit heterogeneous behaviour across lead times.

Future work should extend this framework across stations, pollutants, and climatic regimes, and test whether the ranking reversals observed here persist under broader spatial and temporal heterogeneity. A second priority is to quantify the extent to which apparent predictive gains are reduced when static evaluation is replaced by deployment-realistic rolling-origin evaluation. Such comparisons would help clarify when additional model complexity genuinely extends the operational horizon and when it mainly exploits evaluation artifacts.

\section*{Code Availability}
Code and data will be publicly available at GitHub and Zenodo upon acceptance.

\section*{Data Availability}
The PM10 dataset used in this study is publicly available from the Red de Vigilancia y Control de la Calidad del Aire of the Comunitat Valenciana.

\section*{Software}
All analysis was performed in Python. Code will be released upon acceptance.

\section*{CRediT Authorship Contribution Statement}
Federico Garc\'ia Crespi: Conceptualization, Methodology, Software, Formal analysis, Writing -- Original Draft, Writing -- Review \& Editing.
Eduardo Yubero Funes: Validation, Writing -- Review \& Editing, Supervision.
Marina Alfosea Sim\'on: Validation, Writing -- Review \& Editing, Supervision.

\section*{Declaration of Competing Interests}

The authors declare no competing interests.

\bibliographystyle{elsarticle-num}
\bibliography{references}

@article{EuropeanParliamentandCouncil2008,
  author  = {{European Parliament and Council}},
  title   = {Directive 2008/50/{EC} on ambient air quality and cleaner air for Europe},
  journal = {Official Journal of the European Union},
  year    = {2008},
  volume  = {L 152/1}
}

@article{GomezGomez2025,
  author  = {G{\'o}mez-G{\'o}mez, J. and Guti{\'e}rrez de Rav{\'e}, E. and Jim{\'e}nez-Hornero, F.J.},
  title   = {Assessment of Deep Neural Network Models for Direct and Recursive Multi-Step Prediction of PM10 in Southern Spain},
  journal = {Forecasting},
  year    = {2025},
  volume  = {7},
  number  = {1},
  pages   = {6},
  doi     = {10.3390/forecast7010006},
  url     = {https://doi.org/10.3390/forecast7010006}
}

@article{Krylova2021,
  author  = {Krylova, O. and Okhrin, Y.},
  title   = {Meta-learning framework for forecasting air quality},
  journal = {Environmental Modelling \& Software},
  year    = {2021},
  volume  = {144},
  pages   = {105150},
  doi     = {10.1016/j.envsoft.2021.105150}
}

@article{Vlachogianni2011,
  author  = {Vlachogianni, A. and Kassomenos, P. and Karppinen, A. and Karakitsios, S. and Kukkonen, J.},
  title   = {Evaluation of a multiple regression model for the forecasting of {NOx} and {PM10} in {Athens} and {Helsinki}},
  journal = {Science of the Total Environment},
  year    = {2011},
  volume  = {409},
  pages   = {1559--1571},
  doi     = {10.1016/j.scitotenv.2010.12.040}
}

@article{tashman2000,
  author  = {Tashman, Leonard J.},
  title   = {Out-of-sample tests of forecasting accuracy: an analysis and review},
  journal = {International Journal of Forecasting},
  year    = {2000},
  volume  = {16},
  number  = {4},
  pages   = {437--450},
  doi     = {10.1016/S0169-2070(00)00065-0}
}

@book{hyndman2021,
  author    = {Hyndman, Rob J. and Athanasopoulos, George},
  title     = {Forecasting: Principles and Practice},
  edition   = {3rd},
  publisher = {OTexts},
  address   = {Melbourne, Australia},
  year      = {2021},
  url       = {https://otexts.com/fpp3}
}

@article{cerqueira2020,
  author  = {Cerqueira, Vitor and Torgo, Luis and Mozeti\v{c}, Igor},
  title   = {Evaluating time series forecasting models: an empirical
             study on performance estimation methods},
  journal = {Machine Learning},
  year    = {2020},
  volume  = {109},
  number  = {11},
  pages   = {1997--2028},
  doi     = {10.1007/s10665-020-10076-0}
}

\appendix
\section{Supplementary Material C: \texorpdfstring{H$^*$}{H*} Computation Pseudocode}
\label{app:pseudocode}

Algorithm~\ref{alg:hstar} presents the pseudocode for the H$^*$ computation pipeline. A full Python implementation will be made available upon acceptance.

\begin{figure}[htbp]
\centering
\small
\begin{verbatim}
Algorithm: H* Computation (Rolling-Origin)
INPUT:  Y = {y_1,...,y_T}, model M, max horizon H_max,
        initial train end t_0, evaluation window [t_0+1, T-H_max]
OUTPUT: H*, Skill(h) for h = 1,...,H_max

1. Fix initial training window: Y[1..t_0]
   (e.g., 2017-2019 for daily PM10 series)

2. FOR each monthly origin t in [t_0+1, T-H_max]:
     a. TRAIN:    fit M on Y[1..t-1];
                  estimate preprocessing params P ONLY from Y[1..t-1]
     b. FORECAST: y_hat[t+h] = M(Y[1..t-1], P) for h=1,...,H_max
     c. PERSIST:  pers[t+h] = y_t  (standard lag-1 persistence,
                                     same origin for all horizons)

3. FOR h = 1 to H_max:
     a. RMSE_M[h]    = sqrt( mean_t( (y_hat[t+h] - y[t+h])^2 ) )
     b. RMSE_Pers[h] = sqrt( mean_t( (y_t - y[t+h])^2 ) )
     c. Skill[h]     = 1 - RMSE_M[h] / RMSE_Pers[h]

4. H* = max{ h : Skill[h] > 0 }
   (nominal; interpret together with full Skill[1..H_max] profile)

5. RETURN H*, Skill[1..H_max]
\end{verbatim}
\caption{Pseudocode for H$^*$ computation under rolling-origin validation.
Key constraint: all preprocessing parameters estimated exclusively from
training data at each fold (step~2a).}
\label{alg:hstar}
\end{figure}

\end{document}